\documentclass{article}

\usepackage{arxiv}

\usepackage{enumitem}
\usepackage{lineno}
\usepackage{epstopdf}
\usepackage{float} 
\usepackage{multirow}
\usepackage{longtable}
\usepackage{booktabs}
\usepackage{graphicx} 
\usepackage{graphics} 
\usepackage{caption,subcaption}

\usepackage[colorlinks=true, allcolors=blue]{hyperref}
\newcommand{\email}[1]{\href{mailto:#1}{\tt{\nolinkurl{#1}}}}
\newcommand{\orcid}[1]{ORCID: \href{https://orcid.org/#1}{\tt{\nolinkurl{#1}}}}
\usepackage[sfdefault,lf]{carlito}
\usepackage[parfill]{parskip}

\usepackage{fancyhdr}
\usepackage{authblk}
\usepackage[utf8]{inputenc} 
\usepackage[T1]{fontenc}    
\usepackage{hyperref}       
\usepackage{url}            
\usepackage{booktabs}       
\usepackage{amsfonts}       
\usepackage{nicefrac}       
\usepackage{microtype}      
\usepackage{lipsum}
\usepackage{graphicx}
\usepackage{placeins}
\usepackage{enumitem} 
\graphicspath{ {./images/} }

\title{Real-time Plant Health Assessment via Implementing Cloud-based Scalable Transfer Learning on AWS DeepLens}

\author[1,*]{\textbf Asim Khan}
\author[2]{\textbf Umair Nawaz}
\author[3]{\textbf Anwaar Ulhaq}
\author[4]{\textbf Randall W. Robinson}
\affil[1]{The Institute for Sustainable Industries and \\Liveable Cities (ISILC), Victoria University, Melbourne, Australia. \texttt{asim.khan@vu.edu.au}}
\affil[2]{Namal Institute, Mainwali. \texttt{umair.nawaz@namal.edu.pk}}
\affil[3]{Machine Vision and Digital\\Health Research Group, Charles Sturt University, NSW, Australia. \texttt{aulhaq@csu.edu.au}}
\affil[4]{Applied Ecology Research Group, Victoria University, Melbourne Australia. \texttt{randall.robinson@vu.edu.au}}
\affil[*]{Corresponding author: \email{asim.khan@vu.edu.au}}
\date{}

\begin{document}
\maketitle
\begin{abstract}
In the Agriculture sector, control of plant leaf diseases is crucial as it influences the quality and production of plant species with an impact on the economy of any country. Therefore, automated identification and classification of plant leaf disease at an early stage is essential to reduce economic loss and to conserve the specific species. Previously, to detect and classify plant leaf disease, various Machine Learning models have been proposed; however, they lack usability due to hardware incompatibility, limited scalability and inefficiency in practical usage. Our proposed DeepLens Classification and Detection Model (DCDM) approach deal with such limitations by introducing automated detection and classification of the leaf diseases in fruits (apple, grapes, peach and strawberry) and vegetables (potato and tomato) via scalable transfer learning on AWS SageMaker and importing it on AWS DeepLens for real-time practical usability. Cloud integration provides scalability and ubiquitous access to our approach. Our experiments on extensive image data set of healthy and unhealthy leaves of fruits and vegetables showed an accuracy of 98.78\% with a real-time diagnosis of plant leaves diseases. We used forty thousand images for the training of deep learning model and then evaluated it on ten thousand images. The process of testing an image for disease diagnosis and classification using AWS DeepLens on average took 0.349s, providing disease information to the user in less than a second. 
\end{abstract}



\section{Introduction}
Plant disease has a destructive impact on quantitative and qualitative production \cite{ref1}, leading to a striking blow to producers, traders and consumers. In a US-based study conducted by the U.G.A. Centre for Agribusiness and Economic Development \cite{ref2} discovered a 14.1\% relative disease loss across all crops. A summary of losses due to plant disease included in the 2017 Georgia Farm Gate Value Report (AR-18-01) \cite{ref2} published by University of Georgia Extension. 

Traditionally farmers detect and diagnose plant diseases through their observations and rely upon the opinions of local experts and their past experiences. An expert can decide whether a plant is healthy or not \cite{ref3}. If a plant is found unhealthy, noticeable symptoms on its leaves and fruits are observed and reported. It is hard to correctly diagnose specific diseases even for agronomists and plant pathologists, resulting in incorrect decisions \cite{ref4}. Practical plant health assessment and an early diseases diagnosis can improve product quality and prevent production loss. Early detection and classification of crop disease are significant to secure the specific species production \cite{ref5}. Various research studies have found that early detection of plant diseases is crucial as over the period, diseases start affecting the growth of their species, and their symptoms appear on the leaves \cite{ref6}. When a plant got infected by a specific disease, then significant symptoms are shown on the leaves, which help in the identification and classification of that particular disease \cite{ref7}. Thus controlling and assessment of diseases outspread becomes essential \cite{ref8}. As in peach plant, for instance, the decayed area is small and looks similar in appearance to neighbouring healthy tissue at an early stage; therefore, it is tough to detect diseases \cite{ref2}.

\begin{figure}[H]
	\centering
	\includegraphics[width=13 cm]{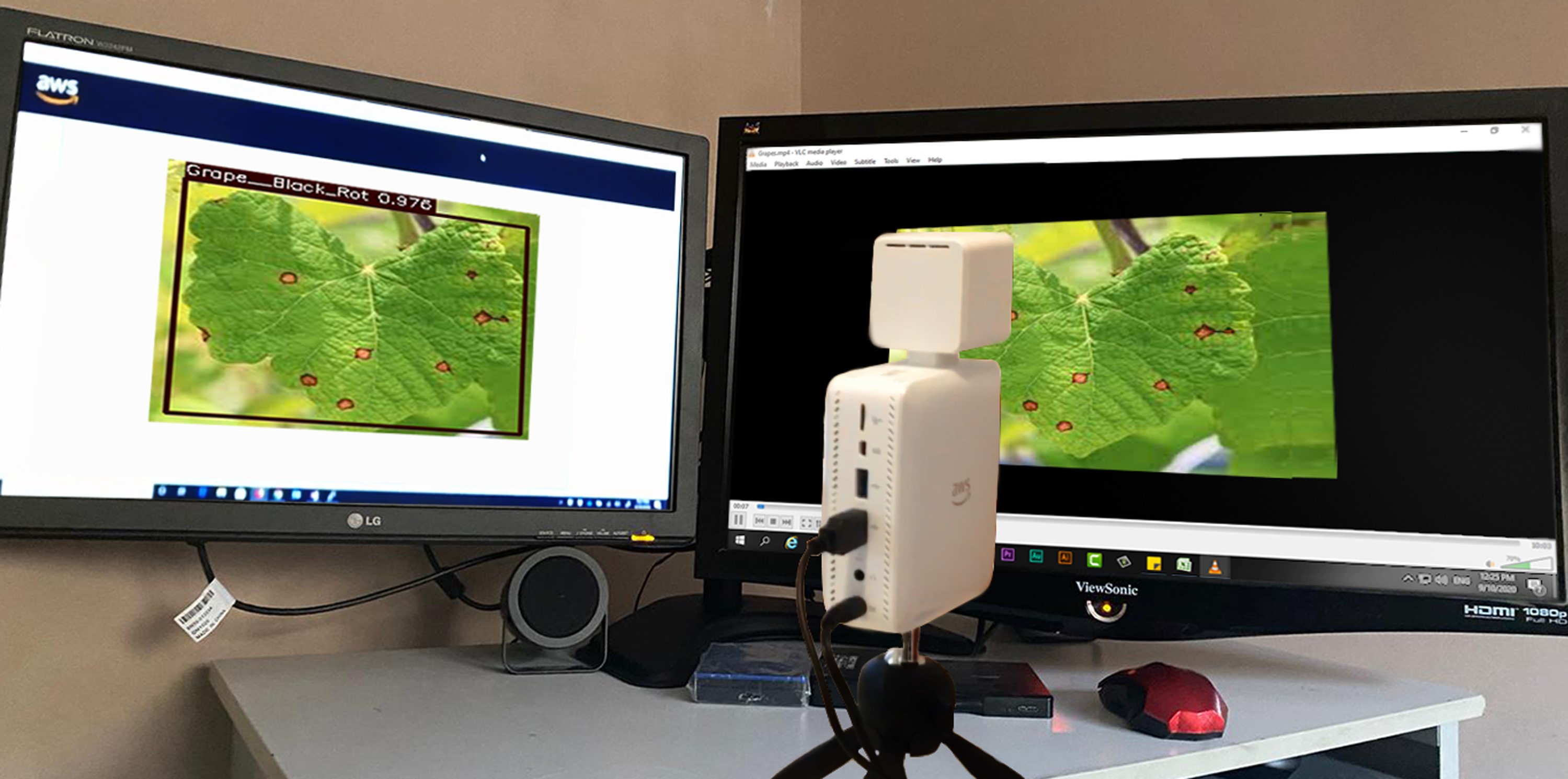}
	\caption{AWS DeepLens Hardware Setup. Input Source (Video) Is In The Right Side Monitor While Output Shown In The Left Side Monitor}
	\label{aws}
\end{figure}
Technology is playing a vital role in exploring agriculture sector. Researchers are trying to explore plant disease detection and classification through the use of different machine learning and image processing techniques. Manual detection of plant diseases is difficult, time-consuming and unreliable. As a health assessment of an individual plant in a large plot is cumbersome and time-consuming, explicitly repeating this checking process over time \cite{ref3}. A single plant may have different diseases having the same pattern of symptoms; moreover, various diseases of the plant show similar signs and symptoms \cite{ref9}, making it challenging to identify the specific disease. Thus technology is helping the agriculture sector, for instance, machine learning (ML) \cite{ref10} algorithms are serving a lot in the process of classification and identification of plant diseases automation. ML helps in monitoring of health assessment of plant and predicting diseases in the plant at early stages \cite{ref7}. With the time progression, new ML models evolved, and the researchers used them for their experiments in the field of recognising and classifying images. Some of those are used in automation of Agriculture systems \cite{ref6}.

For the classification and detection of the plant leaves diseases, several classical and modern ML models are used, such as SVM, VGG architectures, R-FCN, Faster R-CNN, SDD and many others. The advancement in deep learning (DL) \cite{ref11} has provided promising results and solutions in crop disease diagnosis and classification. Islam et al., \cite{ref12} presented the integration of machine learning and image processing for the detection and classification of leaf disease images. They developed an SVM model for potato disease detection and used potato leaves dataset, consisting of healthy leaves and diseased leaves. For performance, they used performance parameters such as accuracy, sensitivity, recall and F1-score. Dubey et al., \cite{ref13} came up with an image processing technique by using the K-Means algorithm for the detection and classification of apple fruit disease and then used multiclass SVM for training and testing images. Al-Amin et al., \cite{ref7} trained their model for potato disease detection through Deep CNN, and they computed performance for analysing the result using parameters such as recall, precision and F1-score.  This model achieved an accuracy of 98.33 \% in experiments. According to Sladojevic et al., \cite{ref14} in order to learn features, CNN must be trained on a large dataset of a large number of images. They developed a CNN model for classification of leaves diseases of apple and tomato plants and the experimental accuracy findings of their research for numerous diseases trial with an accuracy of 96.3\%. Miaomiao et al., \cite{ref15} presented an effective solution for grape diseases detection as they mentioned that two entirely different basic models integrated, it would be more useful to obtain remarkable results and improve the accuracy of detection. Therefore, they proposed a UnitedModel based on the integration of GoogLeNet with ResNet, whereas GoogLeNet raises the total units for all layers of a network and ResNet to raise the total number of layers in a network. Ye Sun et al., \cite{ref16} developed a model based on structured-illumination reflectance imaging (S.I.R.I.) for identification of peach fungal diseases. CNN and three image classification methods used for processing of ratio images, alternating component (AC) images and direct component (DC) images to detect the diseases and area of peach. As a result, they found that A.C. images performance is better than D.C. images in peach diseases detection and ratio images gave a high accuracy rate. Hyeon Park et al.,\cite{ref17} developed a CNN network of two convolutional and three fully connected layers, for disease detection in the strawberry plant. They worked on a small dataset of leaves images consisting of healthy leaves and a powdery mildew strawberries disease class. Xiaoyue et al., \cite{ref18} worked on four typical grapes diseases, and for detection, they proposed a Faster DR-IACNN detector, based on deep learning. They reported that their proposed detector automatically detects the diseased spots on grapes leaves, thus giving an excellent result for the detection of diseases in real-time. In order to detect leaves diseases in vegetables, Zhang et al., \cite{ref19} come up with a model that is RGB colours based three channels CNN.

According to the above-discussed studies, CNN always played a significant role and is widely used in the detection and classification of different plant diseases and provided agreeable results. However, there were some limitations such as these approaches lack usability due to hardware incompatibility issues, limited scalability, inefficiency and some real-time inferences in the practical usage in the real world. Konstantinos et al., \cite{ref4} detected and classified 25 plant diseases by using different CNN based architectures. They trained and tested their model on the open-source dataset named PlantVillage. However, the results obtained in terms of accuracy may differ from using the same dataset for both training and testing purposes. We used a combination of PlantVillage dataset and images collected from the real-cultivation environment and applied different data augmentation techniques on the training data so that we can achieve high accuracy and a robust model.
\begin{figure}[H]
	\centering
	\includegraphics[width=14 cm]{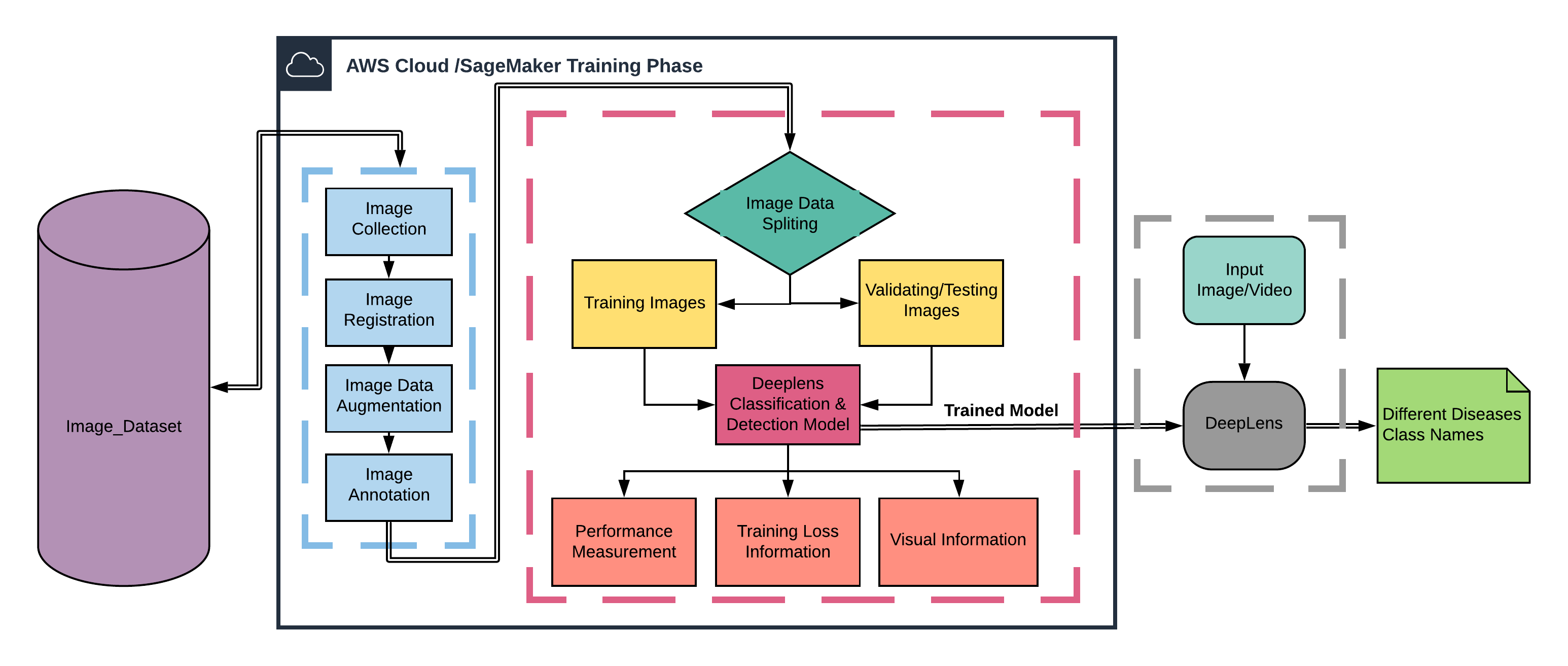}
	\caption{Data Flow Diagram of the DCDM Using AWS Cloud Backend.}
	\label{block}
\end{figure}

Moreover, we used a Cloud-based environment for our training and testing as well as implemented the model in AWS DeepLens for real-time results. The recent development in cloud-based services and efficient deep learning has motivated us to devise practical and scalable solutions to agricultural problems, and this paper lies in the similar domain.  We proposed a model known as DeepLens Classification and Detection Model (DCDM) to detect and classify various fruits and vegetables leaves diseases, based on Deep Convolutional Neural Network (DCNN) with the integration of IoT Device like AWS DeepLens with an average accuracy rate of 98.78\%. In our work, we extracted feature maps \cite{ref14} of an input image after passing through the CNN model and applied filters to visualise the activations through the CNN layers \cite{ref20}. We have trained the DCDM based upon PlantVillage dataset \cite{ref21} and images collected from Tarnab Farm (an agriculture research institute)Pakistan. One limitation we found that most of the images in the PlantVillage dataset are either white or grey background; however, the real-world situation is different and may contain other colours in the background. Thus model trained only uniform background colour may result in low accuracy or wrong prediction. Therefore, in training out DCDM, we included real-world environment field images for training to get maximum accuracy and correct predictions. To make our model scalable and efficient real-time classification and detection, it is integrated with AWS Cloud and implemented in IoT device, namely AWS DeepLens camera. We tested our system on real-time images of twenty-five classes of plant leaves comprising of a healthy leaves class and twenty-four classes of leaves diseases in apple, peach, grapes, strawberry, tomato and potato. The leaves diseases classes for fruits and vegetables that we used in our dataset listed in Table \ref{table1} with both standard and botanical names. Our particular focus is on leaves as they play an essential role in providing energy and producing food supplements for plants. The photosynthesis process produces food in the leaves, and the produced food is supplied to stems and rest of the plant to fulfil its food requirements to keep healthy. Hence, to get more fruit from the plant, their health monitoring is essential. Therefore, our work focuses on early disease detection and classification through health assessment of plant leaves. Another reason for leaf focused disease detection is that leaves remain on the plants for longer times rather than fruits and flowers because they supply energy to the plant.

\section{Materials and Methodology}

The development process of DCDM model for plant leaves disease detection, and classification consists of various stages, i.e. starting with data collection along with data pre-processing and preparation, Training model in AWS Cloud (SageMaker Studio) \cite{ref24} and implementing in AWS DeepLens for inferences purpose, as shown in Figure \ref{aws}. We used real-time videos and images for testing purpose but in the figure we are testing a recorded video on one monitor and output on the other. \\
Further details are as below.

\subsection{Transfer Learning in AWS Cloud}

Transfer learning (TL) is a concept in the ML which simply means that a method learns basic knowledge in solving a particular problem and later reusing that knowledge for other more or less similar problem solution \cite{ref22}. This technique encourages us to use for solving any relevant problem for which there is not sufficient data available. Thus it relaxed the assumption of training and testing data, should be both distributed identically and independently \cite{ref23}. It takes a long time and large-sized dataset for training CNN from the very scratch. Hence in certain situations where the dataset is limited, then TL is a helpful method. For our model, we used TL for different architectures and then training our own model from scratch. To address the scalability limitation, we choose Amazon's Cloud platform and AWS DeepLens. Amazon's cloud platform provides the facility of data storage, data transfer and computational capability. Amazon Web Services (AWS) provides multiple services for many different applications. They also provide a platform to build, train and deploy as well as to validate machine learning models. The trained model can be deployed on AWS Services or any other compatible platforms, for instance, AWS DeepLens. We used SageMaker Studio  for DCDM model training in the AWS Cloud Services so that after completion of training, the obtained trained model could be implanted/deployed in AWS DeepLens as both the platforms are compatible. While comparing with another pre-trained model, we used our system for training and testing purposes.

A typical CNN consists of various layers. Each layer consists of multiple nodes with some activation function attached. The first layer is the input layer that takes input data, whereas, the last layer is the output layer that generates output. A random number of layers exists between input and output layer, referred to as hidden layers (i.e. Convolutional or Convo, Pooling, Dense or Fully Connected and SoftMax layer). If CNN contains two or more than two hidden layers, it is known as Deep Convolutional Neural Network (DCNN) \cite{ref25}.

We designed our DCDM using deep learning TensorFlow framework \cite{ref26} and Keras \cite{ref27} library. Keras is an open-source deep-learning library used to perform different deep learning applications. We used it for the implementation of DCDM architecture, inspired from architecture of VGG19. This architecture uses filters of the same width and height for all the convolutional layers. That is why the architecture of VGG19 used to be very fast than the other state-of-the-art architectures like ResNet50, DenseNet, InceptionVNet. The VGG19 architecture consists of roughly about 139 million parameters \cite{ref28} which makes it computationally expensive for training purpose. However, our architecture has same sequential structure as of VGG19 but with some less number of layers. Also, the numbers of parameters are extensively low, which makes it computationally less expensive. 

\begin{figure}[H]
	\centering
	\includegraphics[width=14 cm]{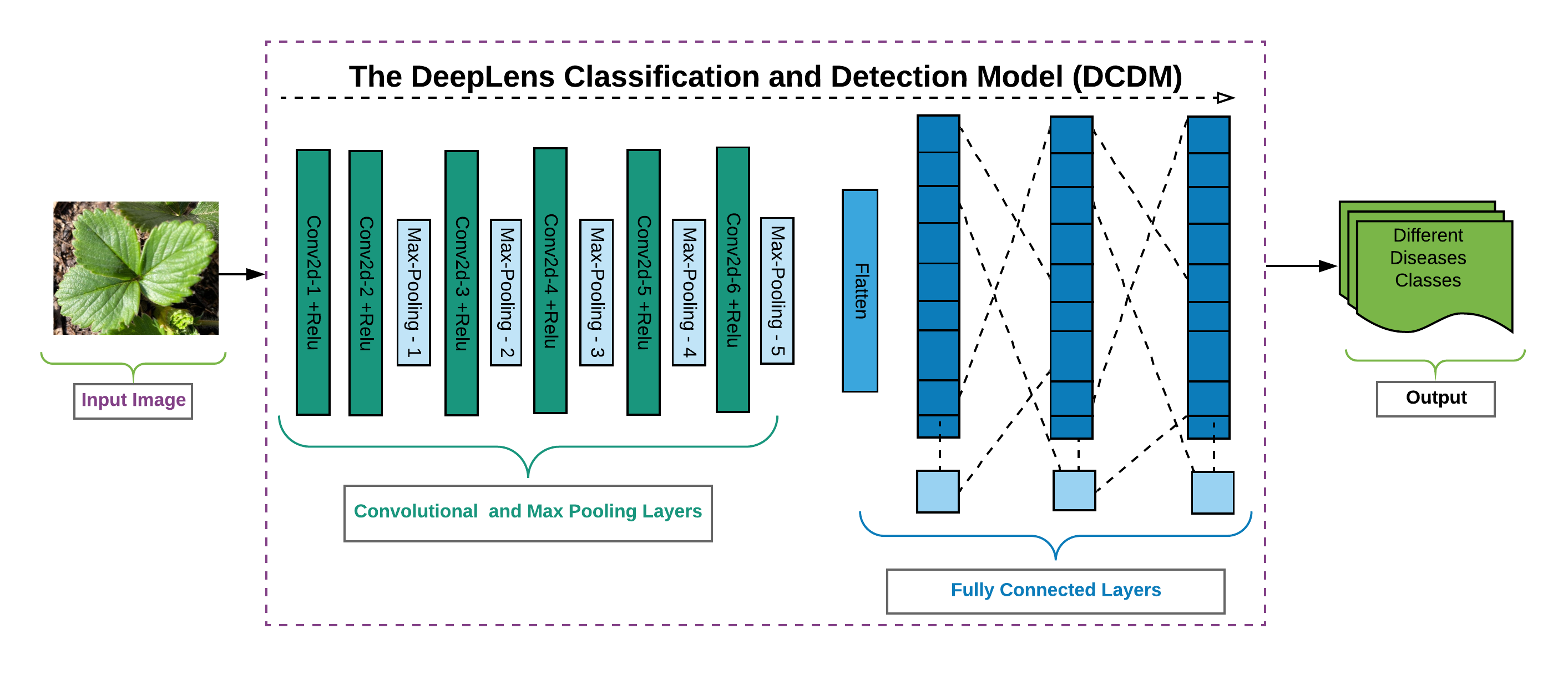}
	\caption{The DCDM Layered Architecture.}
	\label{cnn}
\end{figure}

Our architecture comprises of a total of nine layers with six convolutional layers and three fully connected layers. Figure \ref{cnn} shows visual representation of DCDM Convolutional layers are having non-linearity activation units following by max-pooling layers. The non-linearity activation is often used with convolutional layers. This activation is also known as a ramp function which has a shape of the ramp and transfers the output once it is a positive value; else it results in 0. The last layer, which is also known as a SoftMax layer comprising of 25 nodes is our output layer where each node specifies an individual class of our dataset.\\

The details of these layers are described as:
 
Convolutional Layer: The above stated proposed model used six convolutional layers. There are two types of characteristics in each layer, i.e., input and numeral filters. The filter numbers are then convolved on each layer which extracts the useful features and passes it to the next connected layer. For an RGB image, each filter is applied to all three colour channels, and thus, a corresponding matrix is obtained accordingly. We used a filter size of 3 x 3 for all convolutional layers. The number of filters and input of each layer is elaborated below

\begin{enumerate}[label=(\alph*)]
\item Convolution	272 x 363 – 64 filter
\item Convolution	272 x 363 – 64 filter
\item Convolution	136 x 181 – 128 filte
\item Convolution	68 x 90 – 256 filter
\item Convolution	34 x 45 – 512 filter
\item Convolution	17 x 22 – 512 filter
\end{enumerate}
Pooling Layer: Most commonly, the pooling layer follows each convolutional layer. There are five max-pooling layers in the proposed method. The pooling layers are often used to minimise computational cost as it reduces the size of each convolutional layer output. The max-pooling has an activated filter which slides on the input and based on the size of the filter, and the max value is selected as an output. We used a filter of 2 x 2 for all max-pooling layer. The characteristics for each layer is given below:
\begin{enumerate}[label=(\alph*)]
\item Max-pooling	136 x 181 – 64 filter
\item Max-pooling	68 x 90 – 128 filter
\item Max-pooling	34 x 45 – 256 filter
\item Max-pooling	17 x 22 – 512 filter
\item Max-pooling	8 x 11 – 512 filter 
\end{enumerate}
Dense Layer: It is also known as an artificial neural network (ANN) classifier. Our model has three dense or fully connected layers. In fully-connected layers, each node is connected with only one node of another layer. The first two fully-connected layers have ReLu activation while the last layer, which is also known as the output layer, has a softmax activation. The softmax activation works by finding the node with the highest probability value of prediction being made. Hence the node with the higher value is forwarded as an output.
\begin{enumerate}[label=(\alph*)]
\item Dense Layer: 1024 units\hfill
\item Dense Layer: 1024 units
\item Dense Layer: 25 units
\end{enumerate}

Dropout:  The over-fitting issue is prevented by the addition of a dropout of 0.5. It is added to the dense layers of the model.

Parameters: The total model parameters of our model are 51,161, 305.\\

The model takes the image data as an input, then processes that input data by extracting features from the image and then classifies it either healthy or a diseased leaf, if it is an infected leaf then it further predicts the disease class name, the most resemble one. The predicted class then results as an output.

\subsection{Lambda Function on DeepLens}

\begin{figure}[H]
	\centering
	\includegraphics[width=14 cm]{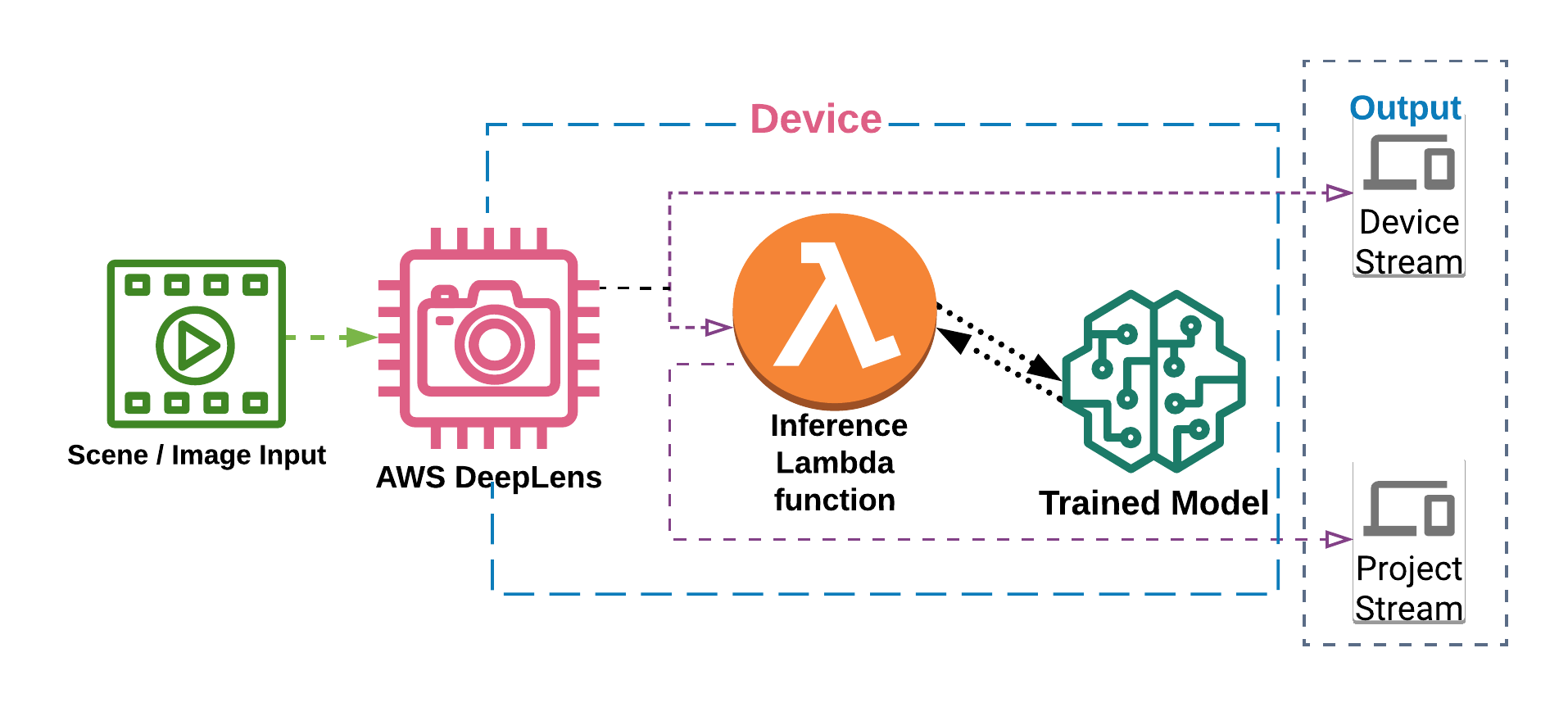}
	\caption{DeepLens [14] Working Framework is Displayed.}
	\label{lamb}
\end{figure}

AWS DeepLens is a deep-learning-based H.D. 4-mega-pixel video camera that is designed specifically for machine learning models developments and implementation. It has a built-in 8GB memory and 16GB storage capacity with 32 GB SD card (extendable). It has more than 100 GFLOPS computing power so it can process machine learning projects independently as well as those integrated with AWS Cloud \cite{ref29}. It has a straightforward usage process as the user can take picture/ image through DeepLens camera, then store it and process it to use in machine learning projects \cite{ref30}. There are a large number of pre-trained models, built-in to it, but a customised model can also be used with DeepLens camera. For instance, any custom based model can be trained or imported into in SageMaker and then can be implemented in AWS DeepLens through various deep learning frameworks such as Tensorflow, Caffe \cite{ref29},\cite{ref30}. A lambda function is used to establish a successful connection to access the DeepLens on a local computer. The lambda functions are the pre-defined functions executed by DeepLens once the project has been deployed \cite{ref31}.  Lambda function streamlines the development process by managing the servers necessary to execute code. They serve as the connection between the AWS DeepLens and Amazon Sagemaker for the camera to generate a real-time inference \cite{ref32}. It controls various resources such as computing capability and power, networking. It has a user-specified function embedded in code, and Lambda function invoke that user code when it is executed. The code returns a message containing data from the event received as input \cite{ref32}. The visual illustration of the AWS DeepLens work-flow in shown in Figure \ref{lamb}.  
After completing the training stage in SageMaker, we implemented the subsequent trained Model in AWS DeepLens camera for inferences of Leaves health assessment. Figure \ref{block}. represents the overall flow of the proposed model.

\begin{longtable}{|l|l|l|l|l|l|l|}
	\caption{The Description Of Leaf Disease Dataset}
	\label{table1}\\
	\hline
	\textbf{\begin{tabular}[c]{@{}l@{}}Class\\    \\ No.\end{tabular}} &
	\textbf{\begin{tabular}[c]{@{}l@{}}Plant \\ Name\end{tabular}} &
	\textbf{\begin{tabular}[c]{@{}l@{}}Plant   \\ Botanical Name\end{tabular}} &
	\textbf{Disease   Name} &
	\textbf{\begin{tabular}[c]{@{}l@{}}Disease   \\ Botanical Name\end{tabular}} &
	\textbf{\begin{tabular}[c]{@{}l@{}}Training \\ Images\end{tabular}} &
	\textbf{\begin{tabular}[c]{@{}l@{}}Validation \\ Images\end{tabular}} \\ \hline
	\endfirsthead
	\multicolumn{7}{c}%
	{{\bfseries Table \thetable\ continued from previous page}} \\
	\hline
	\textbf{\begin{tabular}[c]{@{}l@{}}Class\\    \\ No.\end{tabular}} &
	\textbf{\begin{tabular}[c]{@{}l@{}}Plant \\ Name\end{tabular}} &
	\textbf{\begin{tabular}[c]{@{}l@{}}Plant   \\ Botanical Name\end{tabular}} &
	\textbf{Disease   Name} &
	\textbf{\begin{tabular}[c]{@{}l@{}}Disease   \\ Botanical Name\end{tabular}} &
	\textbf{\begin{tabular}[c]{@{}l@{}}Training \\ Images\end{tabular}} &
	\textbf{\begin{tabular}[c]{@{}l@{}}Validation \\ Images\end{tabular}} \\ \hline
	\endhead
	1 &
	Apple &
	Malus Domestica &
	Scab &
	Venturia   inaequalis &
	1504 &
	326 \\ \hline
	2 &
	Apple &
	Malus Domestica &
	Black rot &
	\begin{tabular}[c]{@{}l@{}}Botryosphaeria   \\ obtusa\end{tabular} &
	1496 &
	325 \\ \hline
	3 &
	Apple &
	Malus Domestica &
	Cedar apple rust &
	\begin{tabular}[c]{@{}l@{}}Gymnosporangium   \\ juniperivirginianae\end{tabular} &
	1220 &
	455 \\ \hline
	4 &
	\begin{tabular}[c]{@{}l@{}}Apple \\ (Healthy)\end{tabular} &
	Malus Domestica &
	&
	&
	1395 &
	329 \\ \hline
	5 &
	Grapes &
	Vitis vinifera &
	Black rot &
	Guignardia   bidwellii &
	1944 &
	236 \\ \hline
	6 &
	Grapes &
	Vitis vinifera &
	Esca &
	\begin{tabular}[c]{@{}l@{}}Phaeomoniella   \\ chlamydospora\end{tabular} &
	1107 &
	276 \\ \hline
	7 &
	Grapes &
	Vitis vinifera &
	Leaf blight &
	\begin{tabular}[c]{@{}l@{}}Pseudocercospora  \\  vitis\end{tabular} &
	1860 &
	215 \\ \hline
	8 &
	\begin{tabular}[c]{@{}l@{}}Grapes\\ (Healthy)\end{tabular} &
	Vitis vinifera &
	&
	&
	1339 &
	484 \\ \hline
	9 &
	Peach &
	Prunus persica &
	Bacterial spot &
	\begin{tabular}[c]{@{}l@{}}Xanthomonas  \\  campestris\end{tabular} &
	1838 &
	459 \\ \hline
	10 &
	\begin{tabular}[c]{@{}l@{}}Peach\\ (Healthy)\end{tabular} &
	Prunus persica &
	&
	&
	1288 &
	572 \\ \hline
	11 &
	Potato &
	Solanum tuberosum &
	Early blight &
	Alternaria   solani &
	1800 &
	200 \\ \hline
	12 &
	Potato &
	Solanum tuberosum &
	Late blight &
	\begin{tabular}[c]{@{}l@{}}Phytophthora   \\ infestans\end{tabular} &
	1800 &
	200 \\ \hline
	13 &
	\begin{tabular}[c]{@{}l@{}}Potato\\ (Healthy)\end{tabular} &
	Solanum tuberosum &
	&
	&
	1121 &
	531 \\ \hline
	14 &
	Strawberry &
	Fragaria spp. &
	Leaf scorch &
	\begin{tabular}[c]{@{}l@{}}Diplocarpon   \\ earlianum\end{tabular} &
	1887 &
	350 \\ \hline
	15 &
	\begin{tabular}[c]{@{}l@{}}Strawberry\\ (Healthy)\end{tabular} &
	Fragaria spp. &
	&
	&
	1364 &
	492 \\ \hline
	16 &
	Tomato &
	\begin{tabular}[c]{@{}l@{}}Lycopersicum \\ esculentum\end{tabular} &
	Bacterial spot &
	\begin{tabular}[c]{@{}l@{}}Xanthomonas \\ campestris pv.\\ Vesicatoria\end{tabular} &
	1710 &
	425 \\ \hline
	17 &
	Tomato &
	\begin{tabular}[c]{@{}l@{}}Lycopersicum \\ esculentum\end{tabular} &
	Early blight &
	\begin{tabular}[c]{@{}l@{}}Alternaria   \\ solani\end{tabular} &
	1800 &
	457 \\ \hline
	18 &
	Tomato &
	\begin{tabular}[c]{@{}l@{}}Lycopersicum \\ esculentum\end{tabular} &
	Late blight &
	\begin{tabular}[c]{@{}l@{}}Phytophthora \\ infestans\end{tabular} &
	1527 &
	382 \\ \hline
	19 &
	Tomato &
	\begin{tabular}[c]{@{}l@{}}Lycopersicum \\ esculentum\end{tabular} &
	Leaf mold &
	Fulvia   fulva &
	1761 &
	491 \\ \hline
	20 &
	Tomato &
	\begin{tabular}[c]{@{}l@{}}Lycopersicum \\ esculentum\end{tabular} &
	Septoria leaf spot &
	Septoria   lycopersici &
	1417 &
	454 \\ \hline
	21 &
	Tomato &
	\begin{tabular}[c]{@{}l@{}}Lycopersicum \\ esculentum\end{tabular} &
	Spider mites &
	Tetranychus   urticae &
	1340 &
	335 \\ \hline
	22 &
	Tomato &
	\begin{tabular}[c]{@{}l@{}}Lycopersicum \\ esculentum\end{tabular} &
	Target spot &
	\begin{tabular}[c]{@{}l@{}}Corynespora   \\ cassiicola\end{tabular} &
	1123 &
	481 \\ \hline
	23 &
	Tomato &
	\begin{tabular}[c]{@{}l@{}}Lycopersicum \\ esculentum\end{tabular} &
	Leaf curl virus &
	&
	3286 &
	571 \\ \hline
	24 &
	Tomato &
	\begin{tabular}[c]{@{}l@{}}Lycopersicum \\ esculentum\end{tabular} &
	Mosaic virus &
	\begin{tabular}[c]{@{}l@{}}Tomato   mosaic \\ virus\end{tabular} &
	1800 &
	574 \\ \hline
	25 &
	\begin{tabular}[c]{@{}l@{}}Tomato\\ (Healthy)\end{tabular} &
	\begin{tabular}[c]{@{}l@{}}Lycopersicum \\ esculentum\end{tabular} &
	&
	&
	1273 &
	380 \\ \hline
\end{longtable}
\FloatBarrier

\section{Dataset Preparation}

\begin{figure}[]
	\centering
	\includegraphics[width=16 cm]{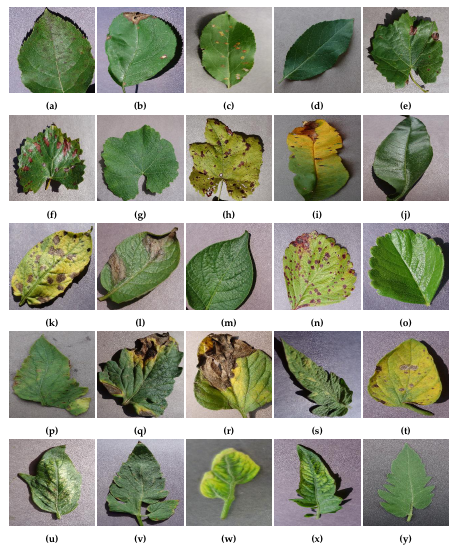}
	\caption{Sample Dataset Images: (a). Apple Scab (b). Black Rot (c). Cedar Apple Rust (d). Apple Healthy (e). Grape Black Rot (f). Grape Esca (g). Grape Leaf Blight (h). Grape Healthy (i). Peach Bacterial Spot (j). Peach Healthy (k). Potato Early Blight (l). Potato Late Blight (m). Potato Healthy (n). Strawberry Leaf Scorch (o). Strawberry Healthy (p). Tomato Bacterial Spot (q). Tomato Early Blight (r). Tomato Late Blight (s). Tomato Leaf Mold (t). Tomato Septoria Leaf Spot (u). Tomato Spider Mites (v). Tomato Target Spot (w). Tomato Leaf Curl Virus (x). Tomato Mosaic Virus (y). Tomato Healty}
	\label{dataset}
\end{figure}

We used a collection of plant leaves images (including both healthy and infectious leaves images for fruits and vegetables) from local farmlands and publicly available dataset known as PlantVillage \cite{ref21}. We analysed around 50,000 images of plant leaves, which categorised into 25 classes and assigned labels to all. Each label represents either a plant disease class or its healthy nature class. A sample image for each class shown in Figure \ref{dataset}.

\FloatBarrier

\subsection{Data Augmentation}

\begin{figure}[H]
	\centering
	\includegraphics[width=16 cm]{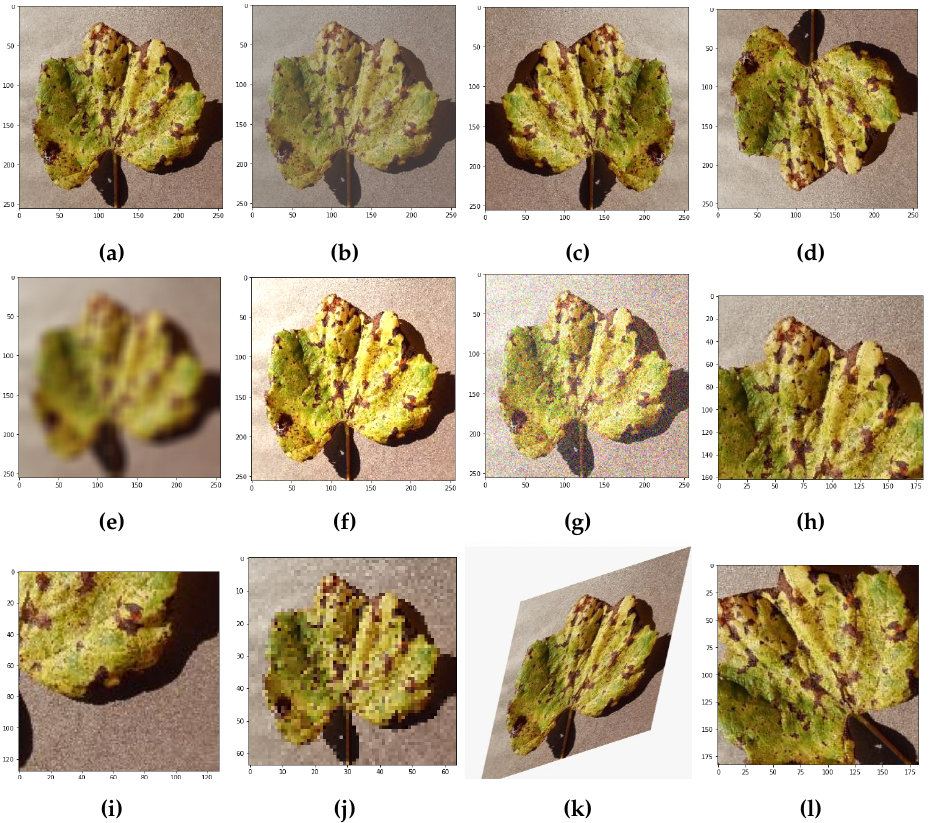}
	\caption{Various Data Augmentation Technique Examples: (a). Original Image (b). Random Contrast (c). Horizontal Flip (d). Vertical Flip (e). Blur (f). Random Bright (g). Random Gaussian Noise (h). Random Crop (i). Deterministic Crop (j). Scale Proportionality (k). Y-Sheared (l) Rotate Without Padding.}
	\label{data_aug}
\end{figure}

For training a DCNN model, a large number of images used for achieving a highly accurate prediction and accuracy. In our case, some of the plants leaves disease classes had fewer images in number; therefore, the process of data augmentation (technique) applied to those limited number of image diseases classes. The process of data augmentation \cite{ref33} provided us with new images from our existing images. Different augmentation techniques like blurriness, rotation, flipping (horizontal and vertical), shearing (horizontal and vertical), and addition of noise were applied accordingly. An illustration of different augmentation techniques shown in Figure \ref{data_aug}. By using this technique, the number of images in our dataset increased, which is essential for obtaining more accurate results after the training stage of CNN. 

The techniques used for augmentation adds new information to the existing images.  For instance, the addition of noise, Gaussian noise is added to the image, and thus an image gets noisy. For horizontal flipping, the image is flipped at the centre on x-axis. Thus, the information from the right side is shifted to the left side. For rotation, a certain angle is applied to the original image, and thus a new image is generated. All these augmentation techniques tweak the present information of an image and help to generate new images with some modifications.

\subsection{Image Registration and Classes Annotation}

After completion of data augmentation process, we had to re-register the images in same dimensions. As we used two types of dataset having different dimensions. Image registration \cite{ref34}\cite{ref35}  is an essential step in the image processing like quality improvement and formation of geometrically aligning images like others in the same dataset. We resized all the images into 272 x 363 pixels and annotated all the images before putting image as an input to any model/network for pre-training CNN structures. The labelled class names listed in Table \ref{table1}.


\subsection{Features Maps Extraction and Filters Visualization in CNN Layers}

\subsubsection{Extraction of Feature Maps}

\begin{figure}[H]
	\centering
	\includegraphics[width=15 cm]{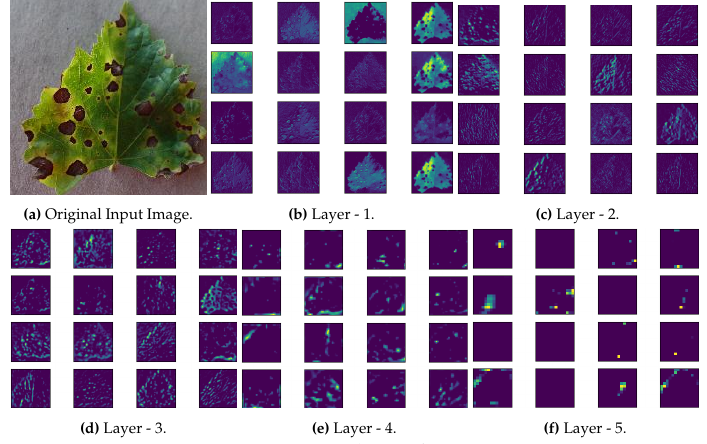}
	\caption{Feature Map Visualisation of DCDM Layers.}
	\label{features}
\end{figure}

Feature maps \cite{ref36} are used to present the local information passing through the CNN Layers. In an ideal feature mapping of CNN, they are sparse and help in the understanding of the classical model. In convolutional layer, to extract feature maps from the source image, several mathematical computations are carried out \cite{ref37}. In Figure \ref{features}, a visual representation for the extraction of feature maps presented for various layers of our model. It also provides information about each layer, i.e. what and how a particular layer of CNN gains information from other layers, such piece of information can help the developer to make proper adjustments in the developing model for best results. From our visualisation images, we found that our model is gaining information in the hierarchical order. It means that the high-level layers present more specific features and vice versa.

Similarly, if the dimensions are higher than feature maps would also classify images more accurately. For instance, in an image, edge corners, some abstract colour features are presented by a deep layer Figure \ref{features}, while other corners and edges represented in shallows layers. Moreover, the middle layers are usually responsible for capturing the same textures because these layers are having complex invariance and more layers in number, after extracting higher-level abstract features, the striking posture of the entire image shown by the high-level feature map.

\FloatBarrier

The feature maps extracted in the first layer represents the overall physical appearance of the leaf image. In the middle layers, the patterns of disease are extracted as can be seen in Figure \ref{features}. The last layers in Figure \ref{features} often extract the delicate features as they are then used to finalise the predicted class.

\subsubsection{Filter Visualization in Model Layers}

\begin{figure}[H]
\centering
\includegraphics[width=15 cm]{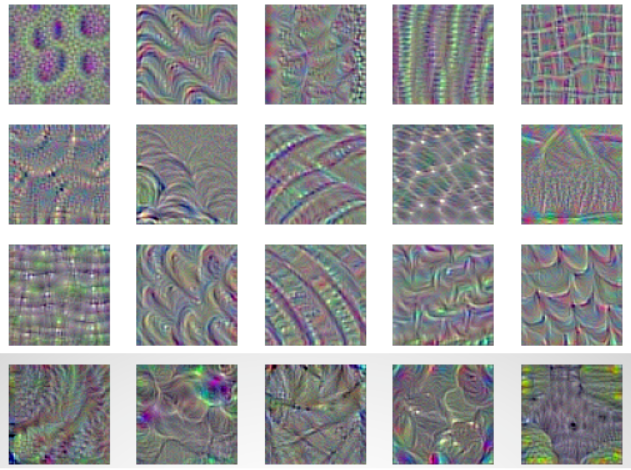}
\caption{Filter Visualisation.}
\label{filters}
\end{figure}

Generally, filters are used for the detection of unique patterns in an input image. It is done by detecting the change in the intensity values of the image. Thus, each filter has its particular importance for feature extraction \cite{ref38}. As an example, a high pass filter detects the existence of edges in an image. In our CNN model, various filters are used to extract features like edges, shape, the colour of the leaf, and many more useful features. In Figure \ref{filters}, a visual representation for a few filters presented where each filter has its application for extracting leaf features. After detecting the specific feature of the image by a filter, it is then passed to the next layer where other filters extract the additional feature. This process continues until the last layer, and thus integrating all together helps to define the predicted class for an input image.
\FloatBarrier



\section{Results and Discussion}

After implementing the trained Model in AWS DeepLens, Experiments for DCDM model done on both Windows and Ubuntu operating systems workstations. It had Intel 9th Gen i7 CPU, i.e. 9700K, 32 GB ECC RAM and NVIDIA RTX TITAN 24 GB VRAM GPU. Table \ref{table2}. shows the system specifications. 

\begin{table}[H]
\caption{System Specification for Testing}
\centering
\begin{tabular}{cc}
\toprule
\textbf{ System Hardware / Software (Operating System) } & \textbf{Specifications}\\
\midrule
CPU			& Intel 9th Gen i7 9700K	\\
RAM			& 32 GB ECC RAM			\\
GPU			& NVIDIA RTX TITAN 24 GB VRAM	\\
Operating System	&Windows 10 Professional and Ubuntu 18.04 \\
\bottomrule
\end{tabular}
\label{table2}
\end{table}

As mentioned above, the deep learning-based framework TensorFlow \cite{ref26} and Keras Library \cite{ref27} were the environments used for experiments.

\begin{figure}[H]
\centering
\includegraphics[width=14 cm]{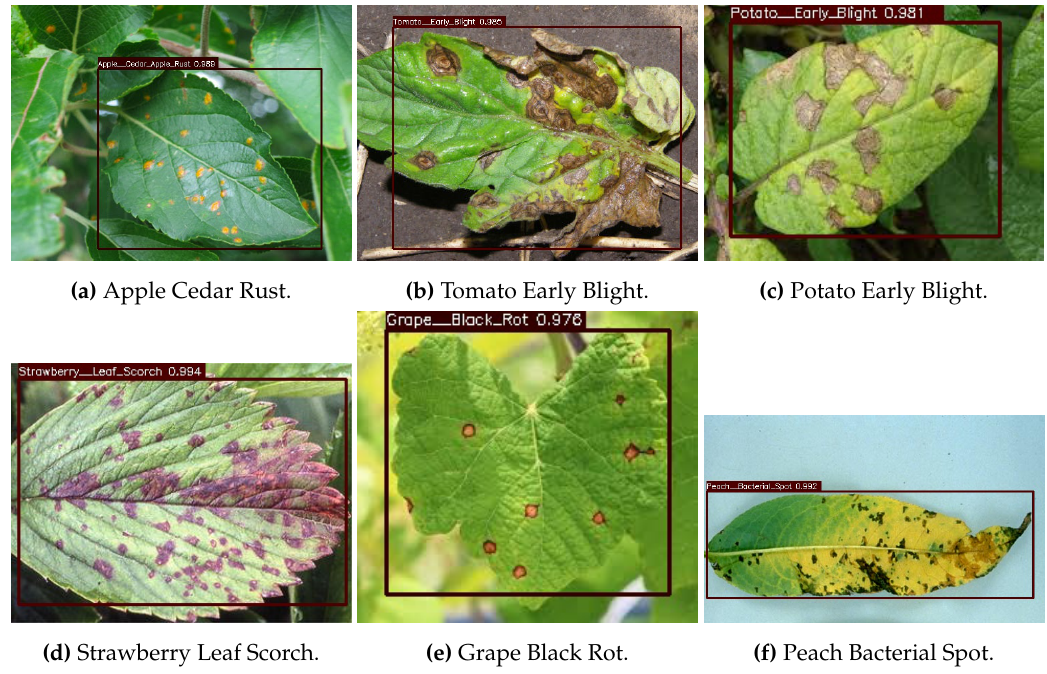}
\caption{Sample Output Images.}
\label{output}
\end{figure}

To measure the performance of our approach and to prevent the issue of over-fitting, we also distributed the data into different training-testing sets.  First, we split the data into 80-20, in which 80\% is used for training whereas 20\% for testing purpose. Again, split into 70-30 means 70\% of the dataset used for training and 30\% for testing purpose and lastly, split into 60-40 training-testing dataset, means 60\% of the complete dataset used for training and 40\% for testing purpose. The total number of each disease class after the augmentation process and the splitting data ratio is given in detail in Table \ref{table3}. PlantVillage dataset contains numerous images of the same plant leaves. During the process of splitting data into a training-testing dataset, we kept all data of the same class in one group, i.e. either in the training set or in the testing set. However, the data split of 80-20 performed very well for DCDM approach with the maximum accuracy of 98.78\%. Thus, exceeding the accuracy results for all other CNN architectures.  Some of the sample output images  with an AWS DeepLens are shown in Figure \ref{output}.

\begin{table}[H]
	\caption{Dataset Split.}
	\centering
	\begin{tabular}{ccc}
		\toprule
		\textbf{Train Test Data Split \%}& \textbf{Training Images}& \textbf{Testing Images}\\
		\midrule
		80 - 20		& 40000			& 10000\\
		70 - 30		& 35000			& 15000\\
		60 - 40		& 30000			& 20000\\
		\bottomrule
	\end{tabular}
	\label{table3}
\end{table}

\subsection{Performance Measurement}

\begin{figure}[H]
\centering
\includegraphics[width=14 cm]{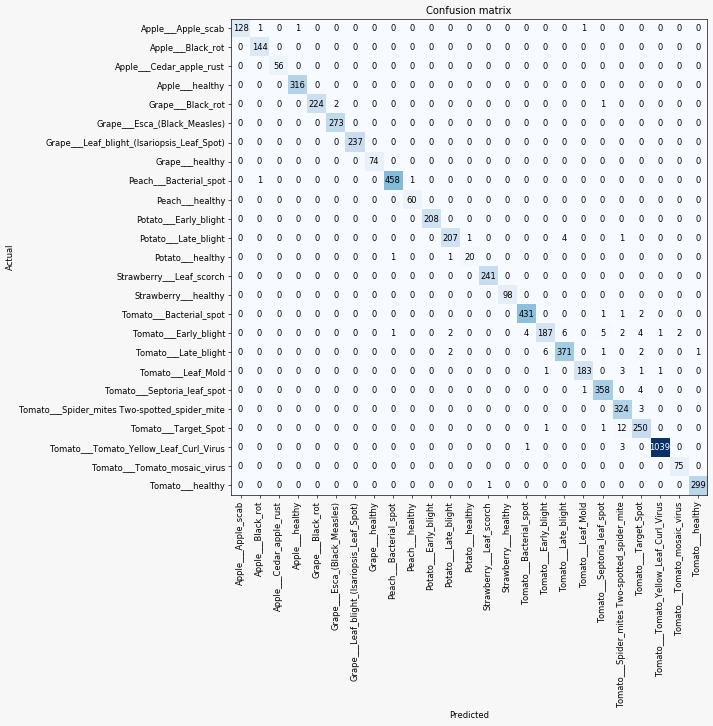}
\caption{Confusuion Matrix for 80 -20 \% Data Split.}
\label{confusion}
\end{figure}

There are many ways for performance measurement that are used to evaluate the performance of neural networks. They include precision, recall, accuracy, and f1-score. The precision tells us about the correct predictions made out of false-positive while recall tells us about the correct predictions made out of false negatives. The accuracy is the number of correct predictions out of both false positives and false negatives. We calculated all the performance measures for our trained model using formulas listed in Eq (1), (2), (3), and (4) from the confusion matrix shown in Figure \ref{confusion}.

\begin{equation}
Precision=\frac{TP}{TP+FP}	
\end{equation}

\begin{equation}
Recall=\frac{TP}{TP+FN}
\end{equation}

\begin{equation}
Accuracy=\frac{TP+TN}{TP+TN+FN+FP}	
\end{equation}

\begin{equation}
F1-Score=2*\: \frac{precision\: *\: recall}{precision+recall}
\end{equation}

Where TP is true positives, TN is true negatives, FP is false positives and FN is false negatives. Here the TP and TN are the correct predictions while the FP and FN are the wrong predictions made by our model. After computing values from the confusion matrix, the results are shown for the 80-20 split in Table 4.\\

\begin{table}[H]
\caption{Classification / Model Performance Report}
\centering
\begin{tabular}{cc}
\toprule
\textbf{ Evaluation Metrics} & \textbf{Value in \%}\\
\midrule
Precision			& 98.38\%	\\
Recall			& 97.98\%			\\
Accuracy			& 98.78\%	\\
F1-Score			& 98.17\% \\
\bottomrule
\end{tabular}
\label{table4}   
\end{table}

In Table \ref{table5}, the performance measure of accuracy for each data split is presented. The values are presented after every ten epochs of training. The bold value for data split of 80-20 and epoch size of 50 represents the highest accuracy.
\begin{longtable}[c]{|c|c|c|c|c|c|}
\caption{Data Split for Testing / Training \& Accuracy Obtained }
\label{table5}\\
\hline
\multicolumn{1}{|l|}{\multirow{2}{*}{\textbf{Dataset (Train/Test) Split in \%}}} &
  \multicolumn{5}{c|}{\textbf{Accuracy {[}\%{]}}} \\ \cline{2-6} 
\multicolumn{1}{|l|}{} &
  \multicolumn{1}{l|}{\textbf{10 Epochs}} &
  \multicolumn{1}{l|}{\textbf{20 Epochs}} &
  \multicolumn{1}{l|}{\textbf{30 Epochs}} &
  \multicolumn{1}{l|}{\textbf{40 Epochs}} &
  \multicolumn{1}{l|}{\textbf{50 Epochs}} \\ \hline
\endfirsthead
\multicolumn{6}{c}%
{{\bfseries Table \thetable\ continued from previous page}} \\
\hline
\multicolumn{1}{|l|}{\multirow{2}{*}{\textbf{Dataset (Train/Test) Split in \%}}} &
  \multicolumn{5}{c|}{\textbf{Accuracy {[}\%{]}}} \\ \cline{2-6} 
\multicolumn{1}{|l|}{} &
  \multicolumn{1}{l|}{\textbf{10 Epochs}} &
  \multicolumn{1}{l|}{\textbf{20 Epochs}} &
  \multicolumn{1}{l|}{\textbf{30 Epochs}} &
  \multicolumn{1}{l|}{\textbf{40 Epochs}} &
  \multicolumn{1}{l|}{\textbf{50 Epochs}} \\ \hline
\endhead
80 – 20 & 92.31 & 95.84 & 96.86 & 97.39 & \textbf{98.78} \\ \hline
70 – 30 & 91.23 & 94.89 & 96.15 & 96.77 & 97.46          \\ \hline
60 – 40 & 90.70 & 94.92 & 95.04 & 95.98 & 96.21          \\ \hline
\end{longtable}
In Figure \ref{confusion}, a confusion matrix is presented for a data split of 80 – 20. The confusion matrix shows the prediction of the model on a visual basis. The values on the diagonal are the correct predictions that are made by a model for the testing dataset. While any value other than diagonal represent the wrong prediction of a model. In this matrix, the healthy images of both fruits and vegetables are mostly predicted correctly.  For example, a grape leaf, all the healthy images of grape leaf were classified correctly and was not mixed with any other leaf class of fruits or vegetables. Similarly, the disease class of grape leaf blight was also successfully predicted with no wrong predictions. 

\begin{figure}[H]
	\centering
	\mbox{}\hfill
	\begin{subfigure}[!h]{0.4\textwidth}
		\centering
		\includegraphics[width = \textwidth]{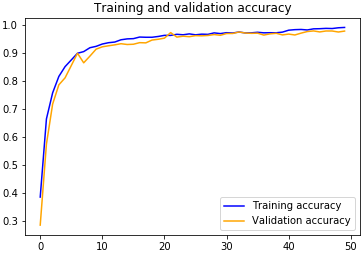}
		\caption{Training and Validation Accuracy}
		\label{figure1}
	\end{subfigure}
	\begin{subfigure}[!h]{0.4\textwidth}
		\centering
		\includegraphics[width = \textwidth]{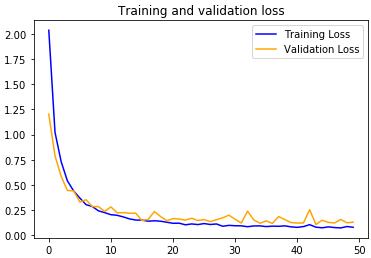}
		\caption{Training and Validation Loss}
		\label{figure2}
	\end{subfigure}
	\hfill\mbox{}
	\caption{Training and Validation Graphs}
	\label{myfigure}
\end{figure}

In Figure \ref{myfigure} (a) (b), the accuracy and loss for both training and testing/validating are presented for each epoch. These graphs were generated for the data split of 80 – 20. The accuracy graph visually shows that accuracy for both training and testing increases gradually and then tends to converge on a specific point. It also shows that after 40 epochs, the change 
in accuracy reduces as the validation accuracy appears to be equivalent to training accuracy.

Similarly, the right graph shows how the loss starts decreasing gradually as the model learns on a given dataset. The loss of validation data becomes stable after 43 epochs and thus tends towards a specific value.

\subsection{Comparative Analysis}
\begin{figure}[!h]
\centering
\includegraphics[width=10cm]{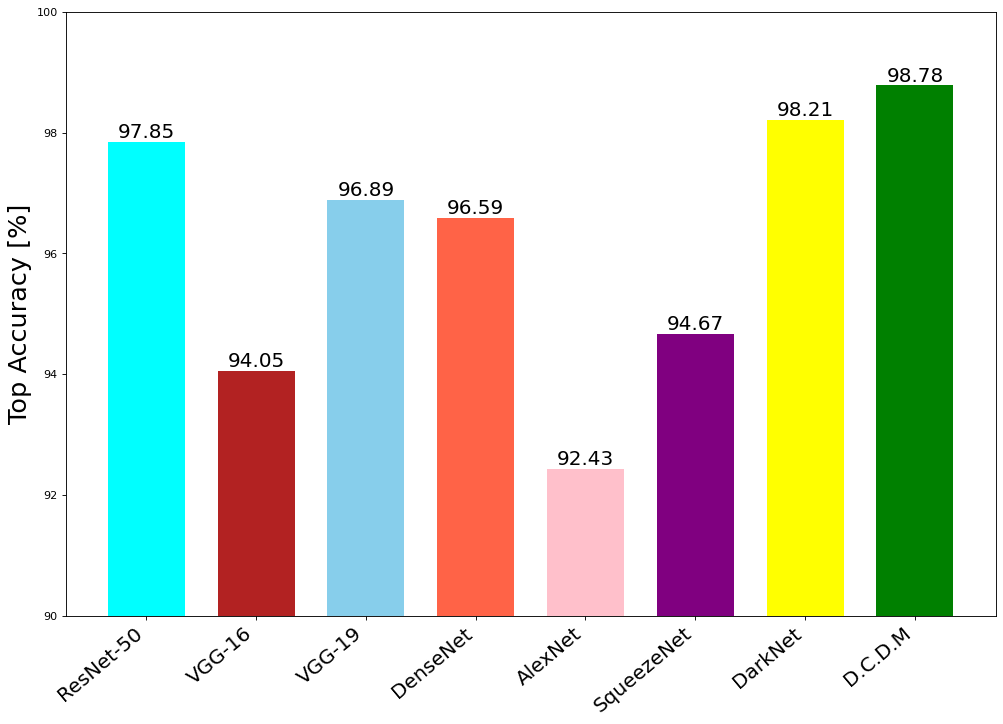}
\caption{Comparison Graph}
\label{cgraph}
\end{figure}

In this section, a visual analysis of different CNN architectures made with the DCDM approach. Training the model on different architecture is a critical approach used to find the best architecture for targeted application. The architectures we used are the best performing architectures for the classification problem. We obtained a performance accuracy of 90+ for each trained architecture. The architecture of AlexNet performs with the lowest accuracy of 92.43\%. This architecture is considered as the smallest architecture and most straightforward architecture out of all. However, still providing us with accuracy above 90\%. The VGG16 and VGG19 architectures are the same with some minor modifications and a different number of layers. They have a significant record of performing very well for the classification problems. For our testing dataset, they provide us with an accuracy of 94.05\% and 96.89\% respectively.
Similarly, the architecture of SqueezeNet and DenseNet also performed with an accuracy of 94.67\% and 96.59\%. The ResNet50 architecture is well-known for good performance on large datasets. It has a bulk of 50 layers with different inter-connections. Thus, performing with an accuracy of 97.85\% and being able to score the position of the third-best architecture in our list. The architecture of DarkNet provides an accuracy close to DCDM approach. It results from an accuracy of 98.21\%, scoring the position of second-best architecture while DCDM architecture performed outstanding and stood with the position of best architecture with an accuracy of 98.78\%. The results for each architecture is shown in Table 6 and then visually represented in Figure \ref{cgraph}. We compared the performance of each architecture for the testing dataset. An evaluation metric of accuracy was used for comparison, based on Equation 3.

The comparison of each architecture with respect to time consumed has also been made which results in the time required for training. The average time for each training epoch is presented in Table \ref{table6} The time consumed by our architecture requires less computation time, thus having lowest average time while training. It testifies that our architecture is the most efficient both  performance as well as computation wise.


\begin{longtable}[c]{lll}

\caption{Disease Classes Accuracy}
\label{table6}\\
\toprule
\textbf  {Trained CNN Models} & \textbf  {Accuracy in \%} &  \textbf  {Average Time Per Epoch(in Minutes)}\\
\endfirsthead
\multicolumn{3}{c}%
{{\bfseries Table \thetable\ continued from previous page}} \\
Trained CNN Models & Acccuracy in \%  & Average Time Per Epoch\\
\endhead
\midrule
ResNet - 50        & 97.85\%  	& 2:03    \\
DensNet            & 96.59\%   	& 2:38 \\
VGG-16              & 94.05\%    	& 1:53 \\
VGG-19              & 96.89 \%   	& 1:59 \\
AlexNet            & 92.43 \%     & 1:44 \\
SqueezeNet       & 94.67\%     & 2:32  \\
DarkNet           & 98.21\%     & 2:13 \\
Our Model          & \textbf{98.78\% } & 1:26      
\end{longtable}

\section{Conclusion}

This proposed deep model implemented on AWS DeepLens can predict 25 different disease classes in Apple, Grape, Peach, Potato, Strawberry and Tomatoes in real-time. Our model obtained 98.78\% accuracy in predictions and classifications in real-time on-field experimentations. This practical approach would facilitate the agriculture-related professionals and community by contributing to the Agri-economy enhancement as the grave problem of plant (leaves) diseases would be easily detected and classified instantly. In addition, this approach is scalable, and we can add more classes of other vegetables and fruit leaves. In our future work, we would integrate our models to different mobile platforms such as iOS, Windows and Android-based Applications to increase its usability. 
Furthermore, new techniques such as Multi-spectral and Hyper-spectral images should also be experimented for detection and classification of plant diseases.

\bibliographystyle{unsrt}  
\bibliography{references}  



\end{document}